\documentclass[conference]{IEEEtran}
\IEEEoverridecommandlockouts
\usepackage{algorithm}
\usepackage{algpseudocode}
\usepackage{lineno}
\usepackage{cite}
\usepackage{amsmath,amssymb,amsfonts}
\usepackage{mathtools, nccmath}
\usepackage[table]{xcolor}
\usepackage{graphicx}
\usepackage{textcomp}
\usepackage{xcolor}
\usepackage{multicol, caption,graphicx}
\usepackage{color,soul}

\usepackage{booktabs}

\usepackage[hyphens]{url}
\usepackage{hyperref}

\usepackage{multirow}
\usepackage{amssymb}
\usepackage{xcolor}
\usepackage{array}
\usepackage{makecell}

\usepackage{pifont} 

\newcommand{\cmark}{\ding{51}} 
\newcommand{\xmark}{\ding{55}} 

\newcommand{\minisection}[1]{\vspace{0.05in}\noindent {\bf #1}}

\begin{document}

\title{AIRES: Accelerating Out-of-Core GCNs via Algorithm-System Co-Design}

\author{%
\IEEEauthorblockN{%
  Shakya Jayakody,\textsuperscript\,
  Youpeng Zhao,\textsuperscript\,
  Jun Wang\textsuperscript\,
  }
  \IEEEauthorblockA{University of Central Florida}%
  \IEEEauthorblockA{Email: \{shakya, youpeng.zhao, jun.wang\}@ucf.edu}%
}

\maketitle

\begin{abstract}
Graph convolutional networks (GCNs) are fundamental in various scientific applications, ranging from biomedical protein-protein interactions (PPI) to large-scale recommendation systems. 
An essential component for modeling graph structures in GCNs is sparse general matrix-matrix multiplication (SpGEMM). 
As the size of graph data continues to scale up, SpGEMMs are often conducted in an out-of-core fashion due to limited GPU memory space in resource-constrained systems. 
Albeit recent efforts that aim to alleviate the memory constraints of out-of-core SpGEMM through either GPU feature caching,  hybrid CPU-GPU memory layout, or performing the computation in sparse format, current systems suffer from both high I/O latency and GPU under-utilization issues.

In this paper, we first identify the problems of existing systems, where sparse format data alignment and memory allocation are the main performance bottlenecks, and propose AIRES, a novel algorithm-system co-design solution to accelerate out-of-core SpGEMM computation for GCNs.
Specifically, from the algorithm angle, AIRES proposes to alleviate the data alignment issues on the block level for matrices in sparse formats and develops a tiling algorithm to facilitate row block-wise alignment.
On the system level, AIRES employs a three-phase dynamic scheduling that features a dual-way data transfer strategy utilizing a tiered memory system: integrating GPU memory, GPU Direct Storage (GDS), and host memory to reduce I/O latency and improve throughput.
Evaluations show that AIRES significantly outperforms the state-of-the-art methods, achieving up to 1.8$\times$ lower latency in real-world graph processing benchmarks.


\end{abstract}

\section{Introduction}

Graph convolutional networks (GCNs) have emerged as a powerful tool for processing structured data, particularly in the field of natural recommendation systems, and biomedical applications~\cite{bongini2021molecular,kipf2016semi,torng2019graph}. 
One of the most critical operations in GCNs is the sparse general matrix multiplication (SpGEMM) for modeling graph structures~\cite{hamilton2017inductive, kipf2016semi,wu2020comprehensive}.
However, the computational demands of this operation can be formidable, especially when dealing with large-scale graphs, motivating the need for efficient acceleration techniques~\cite{hamilton2017inductive}.
The high computation costs of SpGEMM stem largely from the enormous size and inherent sparsity of matrices. 
For instance, in protein-protein interaction (PPI) networks, there are approximately 68 million nodes and 1.5 billion edges, exemplifying such challenges~\cite{hamilton2017inductive, kipf2016semi};
the multiplication and addition operations for each non-zero element could amount to over 1.36 billion.

The advent of high-performance computing and the ever-increasing demand for processing large datasets have propelled the development of more efficient computational methods and hardware accelerators~\cite{owens2008gpu, luebke2008cuda}. 
Among these advancements, GPUs have emerged as a pivotal technology, offering substantial parallel processing capabilities that far exceed those of traditional CPUs. 
Such a technological leap has proven particularly beneficial in matrix operations, which are foundational elements integral to the broader field of machine learning~\cite{nickolls2008scalable, bell2009implementing}.
Sparse matrices, characterized by their predominantly zero-valued elements, present unique challenges and opportunities for optimizing GCN system performance~\cite{alahmadi2020performance, li2020scalable, kipf2016semi}. 
Unlike traditional dense matrix operations, sparse matrices require specialized handling, particularly when scaled to large-scale datasets. 
This necessity becomes critical in out-of-core scenarios where the data could easily exceed in-memory capacities, driving the need for innovative approaches that leverage the raw computational power and memory hierarchy of modern computer systems to maintain or enhance performance without the significant overhead~\cite{saad2003iterative, sun2023helios, waleffe2023mariusgnn}.

Several prior works have attempted to alleviate the memory and computation bottlenecks for large graph-based networks. Notably, UCG~\cite{lin2024unified} proposes a unified CPU-GPU protocol that dynamically balances the workload between the CPU and GPU, incorporates GPU feature caching, unified shared memory, and overlaps communication with computation.
ETC~\cite{gao2024etc} further introduces a new batching scheme with a specialized data access strategy for matrices in sparse formats to avoid redundant data transfers.
However, despite their promising results, today's GCN systems continue to struggle to achieve high end-to-end throughput, especially in out-of-core environments with a limited GPU memory capacity.

Sparse matrices are often stored in compressed formats to reduce memory usage. In SpGEMM, one common method is to decode the compressed matrix into a dense format before performing the multiplication. Unfortunately, it incurs significant decoding latency and imposes large GPU memory space. An alternative strategy is to perform SpGEMM directly on the compressed (sparse) data on the fly, thereby reducing both memory usage and decoding latency. Nevertheless, even in the compressed form, matrices derived from large, intensive graph data can be extremely large, potentially exceeding available GPU memory and triggering out-of-core processing. To conduct a practical out-of-core SpGEMM with compressed formats, it poses new challenges:
1) \textbf{lack of data alignment control}, 
and 2) \textbf{inflexible static memory allocation strategy}.
More specifically, large sparse data is generally transferred incrementally by a chunk of data called segments from host memory to the GPU.
One straightforward way is to transfer as many rows/columns to maximize GPU memory usage. 
However, we observe that such a method induces significant CPU pre-processing and I/O overheads:
after each compute cycle, the next portion of data often needs to be packed with the last portion of data already transferred to the GPU for merging and staging in the host memory. 
This is because, upon receiving a non-integer number of rows, the GPU performs matrix operations on complete row data while returning incomplete row data back to the CPU host. 
Moreover, different from computing in dense formats, the memory required for output allocation is usually dynamic instead of static, decided by both the matching process between the compressed format of the first matrix row and the second matrix column. 
As the exact number of matches cannot be pre-determined, the memory allocation remains unknown beforehand.
Therefore, flexible and adaptive memory management is needed to avoid GPU resource under-utilization and lengthy execution times.
Addressing these challenges is essential for enhancing the capabilities of out-of-core GCN systems, ensuring they are equipped to efficiently manage the growing size and complexity of graph data in diverse real-world applications.

\begin{table}[!t]
\centering
\caption{Comparison of prior works and AIRES.
DMA and UM denote direct memory access and unified memory, respectively.}
\vspace{-1mm}
\label{tab:Evaluation}
\resizebox{\linewidth}{!}{
\begin{tabular}{c||ccc}
\toprule
& \textbf{ UCG~\cite{lin2024unified}} & \textbf{ETC~\cite{gao2024etc}}  &\textbf{AIRES (Ours)} \\ 
\midrule \midrule
Alignment  & \xmark & \xmark   & \cmark\\ 
\midrule
DMA  & \xmark & \cmark &\cmark \\ 
\midrule
UM reads  & \cmark& \xmark  & \xmark \\ 
\midrule
\begin{tabular}[c]{@{}c@{}}Dual-way\end{tabular}  & \xmark& \xmark  & \cmark \\ 
\midrule \midrule
Co-Design & \xmark & \xmark & \cmark \\ 
\bottomrule
\end{tabular}
}
\end{table}

In this paper, we introduce AIRES, a novel algorithm-system co-design approach to tackle the above-mentioned challenges. 
Our focus is on accelerating SpGEMM for GCNs represented in sparse formats, optimized for out-of-core processing on GPUs. 
Specifically, on the algorithm level, AIRES incorporates a block-wise data alignment solution to alleviate data staging overheads and expedite computation. 
Built upon such a data alignment strategy, AIRES features a specialized tiling for block-wise partitioned data to optimize compressed matrix multiplication;
on the systems level, AIRES employs a three-phase dynamic scheduling protocol that leverages a dual-way path strategy 
to perform dynamic output memory allocation to maximize GPU resources and further reduce I/O latency.

We implement and validate ARIES across numerous workload scenarios. 
Extensive evaluations demonstrate that ARIES outperforms
existing GCN systems in both end-to-end latency and overall throughput. 
Specifically, ARIES sustains up to 1.8$\times$ and 1.5$\times$ runtime
improvement against the state-of-the-art solutions,  UCG\cite{lin2024unified}, ETC~\cite{gao2024etc} respectively, across various graph datasets.

In summary, this paper makes the following contributions: 

\begin{itemize}
\item  We identify the data alignment and memory allocation challenges in prior GCN systems and propose an algorithm-system co-design solution, AIRES, for efficient out-of-core GCN training in memory-constrained systems. 

\item On the algorithm level, AIRES features a new block-wise partitioning tailored for compressed matrix formats and a specialized tiling optimization strategy for block-wise partitioned data. 

\item On the system level, we develop a dynamic scheduling strategy that enables efficient data transfer across a tiered memory system using a dual-path data transfer strategy to avoid resource under-utilization.

\item We evaluate AIRES over diverse graph workloads.
Experiments demonstrate that AIRES can significantly reduce end-to-end latency against existing systems.  
\end{itemize}

\section{Background}
\label{sec:bg}


\begin{figure}[!t]
\centerline{
\includegraphics[width=0.45\textwidth]{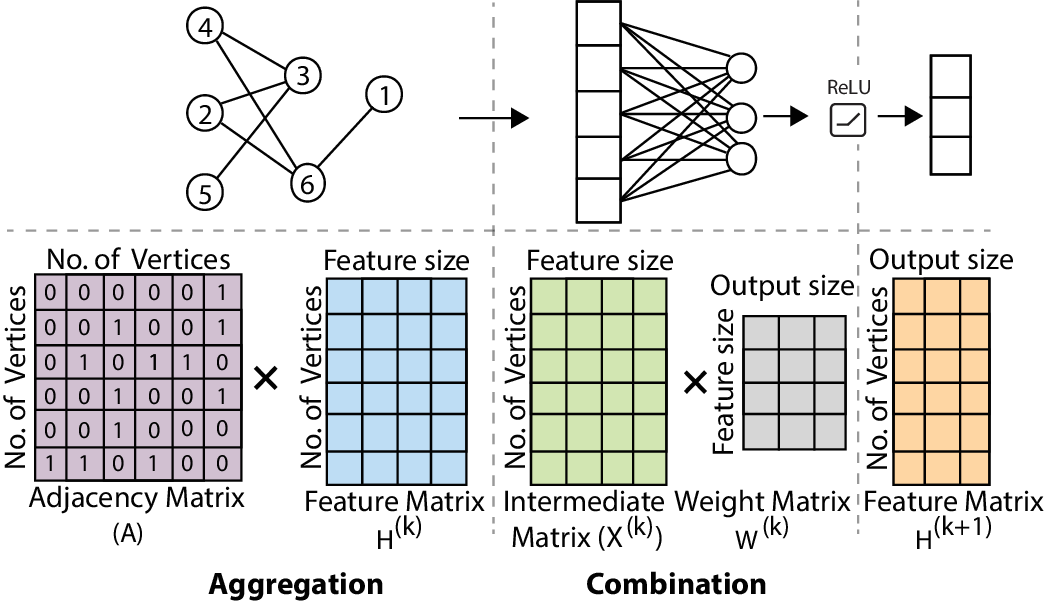}
}
\caption{Illustration of chain matrix multiplication in graph convolutional networks (GCNs) during the aggregation and combination phases.
}
\label{fig:chained_MM}
\end{figure}

\subsection{Graph Convolutional Networks (GCNs)}
Graph convolutional networks (GCNs) exhibit parallels to traditional convolutions in the realm of graph analysis, particularly in their approach of sharing ``filter" parameters across all graph locations~\cite{kipf2016semi, defferrard2016convolutional}. 
Figure~\ref{fig:chained_MM} shows the chain matrix multiplication during the GCN computation. 
The framework of GCN can be broken down into two primary stages:

\begin{figure}[!t]

\centerline{\includegraphics[width=0.4\textwidth]{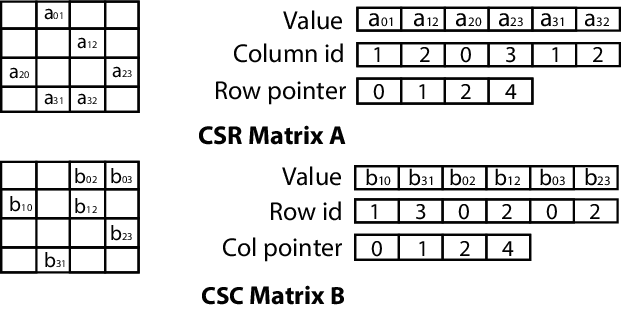}}
\vspace{-1mm}
\caption{Examples of the compressed format of matrix A (CSR) and matrix B (CSC).
}
\label{fig:csr_matrixA_matrixB}
\end{figure}

\minisection{1) Aggregation}: Each node constructs a feature vector, representing the message it intends to disseminate to all its adjacent nodes.
    The aggregation equation is defined as:
    \begin{align}
        X^{(k)} = \tilde{A} H^{(k)} \\
        \tilde{A} = \hat{D}^{-\frac{1}{2}} \hat{A} \hat{D}^{-\frac{1}{2}} 
    \end{align}
    where $\tilde{A}$ and $\hat{A}$ are the normalized and augmented adjacency matrix, respectively, where \(\hat{A} = A + I\), with \(A\) being the original adjacency matrix and \(I\) being the identity matrix, which includes self-loops. The degree matrix \(\hat{D}\) is a diagonal matrix where each diagonal element corresponds to the degree of the respective node, ensuring that the feature aggregation is normalized by the connectivity of each node~\cite{velivckovic2017graph}.
    
\minisection{2) Combination}: These messages are sent to neighboring nodes. Consequently, each node receives one message from every adjacent node.
The combination equation with the transformation through weight parameters and an activation function is given by:
\begin{equation}
H^{(k+1)} = \sigma\left(X^{(k)} W^{(k)}\right)
\end{equation}
where \( W^{(k)} \) represents the weight matrix at layer \( k \) and \( \sigma \) is the activation function, such as ReLU\cite{ kipf2016semi}. This step learns trainable parameters that allow the network to combine the aggregated features effectively, adjusting how each node processes its local neighborhood. 

\noindent Altogether, the computation within a standard GCN layer can be expressed as follows.

\begin{equation}
    H^{(k+1)} = \sigma\left(\hat{D}^{-\frac{1}{2}} \hat{A} \hat{D}^{-\frac{1}{2}} H^{(k)} W^{(k)}\right)
\end{equation}

\subsection{Compressed Matrix Formats}
In the landscape of data-intensive computing applications, the multiplication of large sparse matrices is a fundamental operation that often presents significant computational challenges ~\cite{li2020scalable, ashari2014fast}.
Traditional dense matrix multiplication methods can be highly inefficient when applied to sparse matrices due to unnecessary processing of zero elements, leading to increased computational time and resource consumption~\cite{baskaran2009optimizing}.
Compressed matrix formats, such as compressed sparse row (CSR) and compressed sparse column (CSC), have been developed to address these inefficiencies~\cite{greathouse2014efficient,filippone2017sparse}. 
These formats significantly reduce the storage and computation overhead by only storing and processing non-zero elements of the matrices
An illustrative example of CSR and CSC matrices is shown in Figure~\ref{fig:csr_matrixA_matrixB}.

\section{AIRES Design}
\label{sec:algo}
\subsection{Algorithm Design}

\begin{figure}[!t]
\centerline{\includegraphics[width=0.45\textwidth]{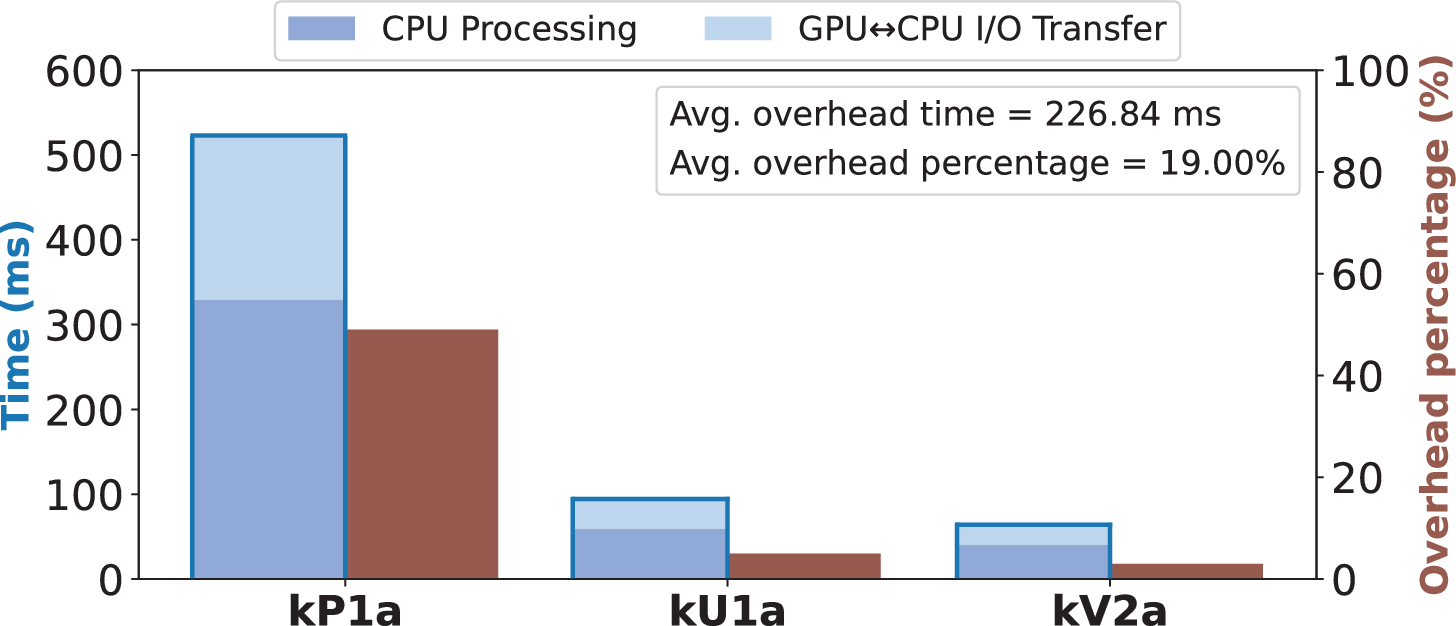}}
\caption{Merging overhead observed across three different datasets.  
The latency overhead includes merging the partial segments, and data transfer time between the GPU and host memory.
The percentage is measured over the computation latency.
}
\label{fig:data_aligment_in_ms}
\end{figure}

\minisection{Row block-wise (RoBW) Data Alignment.}
In the era of big data, existing works mainly aim to address the memory and computation challenges through GPU-CPU coordination~\cite{lin2024unified} and batching schemes for graph data~\cite{gao2024etc} to improve system throughput.
As the GCN computations are often conducted in an out-of-core fashion, large matrices must be transferred in segments to circumvent GPU out-of-memory (OOM) errors.
A naive way to maximize the available GPU memory space is to send out as many rows or columns as possible.
Nevertheless, we observe that segments of the original matrices often contain partial rows/columns, which cannot be processed at the current computation cycle.
Therefore, such partial data must be repetitively transferred back to host memory to merge with the remaining data, which could lead to significant processing latency overheads.

\begin{algorithm*}[!t]
\caption{Pseudo Code for RoBW Partitioning for CSR A.}
\label{algo:csr_block_allocation}
\begin{algorithmic}[1]
\Statex \textbf{Input:} $A^{rp}, A^{col}, A^{v}$: CSR row pointers, column indices, and values for matrix $A$. $N$: Total number of row pointers in matrix  $A$. $M_A$: Available GPU memory for CSR A.
\Statex \textbf{Output:} RoBW segment data $RoBW_p^S=(A_p^{rp}, A_p^{col}, A_p^{v})$ $(p=1,2,...,n)$. 

\State $startIdx = 0, n = 0$

\While{$startIdx < N$} 

    \State $endIdx = startIdx, z = 0$ \Comment{Initialize non-zero count in the block}
    \State $ q = A^{rp}[endIdx + 1] - A^{rp}[endIdx]$\Comment{Non-zero count for the current row}

    \State $ k = endIdx - startIdx + 1$\Comment{Number of rows currently in the block}
    \While{$endIdx < N$ \textbf{and} \texttt{calcMem}($k, q$) $\leq M_A$}\Comment{Add rows to the block until memory exceeds limit}
        \State $z \mathrel{+}= A^{rp}[endIdx + 1] - A^{rp}[endIdx], endIdx \mathrel{+}= 1$ \Comment{Update non-zero count \& expand block}
        \State $ q = z + A^{rp}[endIdx + 1] - A^{rp}[endIdx], k = endIdx - startIdx + 1$

    \EndWhile

    \State \texttt{malloc:} $A_p^{rp}[k], A_p^{col}[z], A_p^{v}[z]$\Comment{Allocate CPU memory}
    
    \State $A_p^{rp}[0] = 0, z_p = 0$ \Comment{Initialize row pointer and non-zero count for the block}
    
    \For{$i \gets startIdx$ \textbf{to} $endIdx - 1$} \Comment{Iterate over rows in block}
        \State $rowstartIdx, rowEndIdx = A^{rp}[i], A^{rp}[i + 1]$ 
        \For{$j \gets rowstartIdx$ \textbf{to} $rowEndIdx - 1$} \Comment{Copy column indices and values}
            \State $A_p^{col}[z_p], A_p^{v}[z_p] = A^{col}[j], A^{v}[j]$
            \State $z_p \mathrel{+}= 1$
        \EndFor
        
        \State $A_p^{rp}[i - startIdx + 1] = z_p$ \Comment{Update row pointer for the block}
    \EndFor
    \State $n \mathrel{+}= 1, startIdx = endIdx$ \Comment{Update block count \& move to the next block}
\EndWhile
\end{algorithmic}
\end{algorithm*}

To corroborate our intuition, we conduct exploratory experiments for three real-world graph datasets, namely kmer\_P1a (kP1a), kmer\_U1a (kU1a), and kmer\_V2a (kV2a) from SuiteSparse~\cite{davis2011university} and showcase the overheads of merging partial data in Figure~\ref{fig:data_aligment_in_ms}.
We have two key observations. 
\underline{First}, the merging overheads are non-negligible. 
For instance, in the kV2a dataset, the merging overhead accounts for a whopping 50\% of the total computation latency.
\underline{Second}, the smaller the allocated GPU memory, the higher the overheads, e.g. the latency overhead of kV2a is about 6$\times$ higher than that of kP1a. 

\begin{figure}[!t]
\centerline{\includegraphics[width=0.45\textwidth]{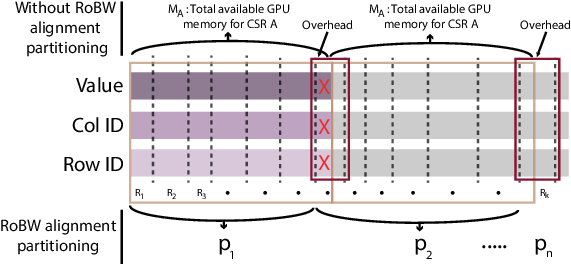}}
\caption{Illustration of row block-wise (RoBW) partitioning for CSR A. 
The top is without RoBW alignment partitioning and the bottom is RoBW alignment partitioning. 
\(R_1\) to \(R_k\) is the matrix rows. $p$ is the RoBW alignment
partitioning size, $n$ is the number of RoBW.
}
\label{fig:Row_block-wise}
\end{figure}

To effectively alleviate the above overhead, we propose performing alignment for sparse data segments.
Specifically, we employ a row block-wise (RoBW) partitioning strategy to avoid unnecessary merging and I/O operations.
Figure~\ref{fig:Row_block-wise} illustrates our RoBW partitioning method for CSR A. 
Instead of naively maximizing the available memory space, we divide the sparse data into blocks, each containing complete and unfragmented data.

The block size $p$ is decided by the dataset on the fly, as each data segment depends on the input CSC B matrix and output matrix CSR C. We first approximate the GPU memory usage $M_C$ for output matrix CSR C as follows:
\begin{align}
\centering
M_C &= \frac{3\cdot\alpha_{A}\cdot(100 - s_{A})}{100} \cdot \left(1 +\frac{\alpha_{B}}{\alpha_{A}}+ \frac{100-s_{B}}{100}\right)
\end{align}
where $\alpha_{A}$ and $\alpha_{B}$ is the value size for CSR A and CSC B, $s_{A}$ is the sparsity percentage for CSR A, and $s_{B}$ is the sparsity percentage for CSC B. 
The memory for CSC B can be calculated as:
\begin{equation}
M_B = (\alpha_B +\beta_B + \theta_B)
\end{equation}
where $\alpha_B, \beta_B, \theta_B$ are the value size,  column id, and row id for CSC B, respectively. 
Therefore, the optimal block size $p$ can be estimated as follows:

\begin{equation}
p = \left(\frac{M - M_C - M_B}{3}\right)
\end{equation}

\begin{figure*}[!t]
\centerline{\includegraphics[width=0.98\textwidth]{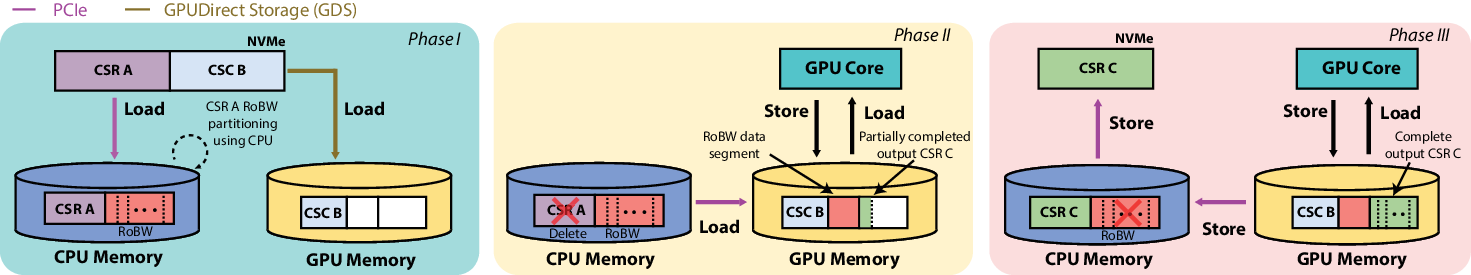}}
\caption{AIRES memory allocation strategy. 
In Phase I, CSA A and CSC B dual-way data transfer to CPU memory and GPU memory, respectively, and CSR A RoBW partitioning using the CPU. 
In Phase II, free memory for CSR A, transfer the RoBW data segment to GPU memory, and compute partial CSR C results using the transferred RoBW data segment. From Phase II to Phase III, free the current RoBW data segment in GPU memory and update it with the next RoBW data segment from CPU to GPU memory, and compute the next partial CSR C; this will keep repeating. In Phase III, complete the output CSR C, free memory for RoBW in CPU memory, transfer the CSR C result to CPU memory, and store the final copy in NVMe.
}
\label{fig:peridot_utilization}
\end{figure*}

Algorithm~\ref{algo:csr_block_allocation} presents details of dynamically adjusting the block size based on the available memory and processing requirements, where we ensure the optimal utilization of GPU resources.  


\subsection{System Design}
In dense matrix multiplication, the amount of memory needed to store the output matrix can be easily determined by examining the row or column size directly.
In the case of sparse formats, the memory allocation varies, as it also depends on the matching process of the first matrix row and the second matrix column.
Therefore, simply reserving the memory based on the matrix size would often lead to resource under-utilization, thus inducing potential latency overheads.
Prior work methods, UCG~\cite{lin2024unified} directly utilize row or column sizes and ETC~\cite{wang2023hongtu} allocate memory equivalent to the larger compressed format, frequently resulting in GPU memory inefficiencies.
To combat this issue, we propose a three-phase dynamic scheduling strategy to allocate GPU memory adaptively.
Specifically, our design features a dual-way data transfer protocol that leverages three tiers of the memory hierarchy, namely GPU memory, host memory, and NVMe secondary storage.

We illustrate our scheduling with an illustrative example in Figure~\ref{fig:peridot_utilization} and a formulation in Algorithm~\ref{algo:scheduling}.
It employs a dual-path strategy where CSC B data is transferred via GPU Direct Storage (GDS) from NVMe directly to GPU memory, while CSR A data is initially transferred to CPU memory where it undergoes preprocessing before being forwarded to the GPU memory. 
This method effectively harnesses the strengths of both GDS and conventional memory pathways, optimizing throughput and reducing delays.
Our scheduling is described as below:

\begin{algorithm}[!t]
\caption{AIRES Dynamic Scheduling.}
\label{algo:scheduling}
\begin{algorithmic}[1]
\Statex \textbf{Initialization:} GPU core $GC$, CPU core $CC$,  GPU memory $GM$, CPU memory $CM$, NVMe memory $NM$, $CSR A$ matrix , $CSC B$ matrix, RoBW segment data $RoBW_p^S$ $(p=1,2,...,n)$.

\State \textcolor{gray}{\# Load}
\State $\text{$NM.CSC B$} \rightarrow \text{$GM$}.\text{load}(\ast)$ \Comment{Phase I}
\State $\text{$NM.CSR A$} \rightarrow \text{$CM$}.\text{load}(\ast)$
\State \textcolor{gray}{\# Compute}
\State $\text{$CC.CSR A$} \rightarrow \text{$CC.RoBW$}.\text{Process}(\ast)$
\For{all $p < n$} \Comment{Phase II}
 \State \textcolor{gray}{\# Load}  
 \If{$GC.CSRA\geq RoBW_p^S $}
 \State $\text{$CM.RoBW_p^S$} \rightarrow \text{$GM.CSRA$}.\text{load}(\ast)$
    \EndIf
   \State $\text{$GM$} \rightarrow \text{$GC$}.\text{load}(\ast)$

 \State \textcolor{gray}{\# Compute} 
 \State $\text{$GC$.Compute}(\ast)$
  \State \textcolor{gray}{\# Store}  
  \State $\text{$GC.CSRC_p$} \rightarrow \text{$GM.CSRC_p$}.\text{store}(\ast)$ 
  \EndFor
  \State \textcolor{gray}{\# Store}
  \State $\text{$GM.CSRC$} \rightarrow \text{$CM.CSRC$}.\text{store}(\ast)$ \Comment{Phase III}

\end{algorithmic}
\end{algorithm}

\begin{itemize}
\item \textbf{Phase I}: All CSC B data will be loaded into the GPU memory, while the CSR A data will be partitioned and stored in the CPU memory using the RoBW strategy. 
\item \textbf{Phase II}: A batch of data segments from the RoBW partition is transferred from the CPU to the GPU memory. Once in the GPU, partial computations are performed on the available data to produce the output (CSR C), which is stored in the GPU memory. 
\item \textbf{Phase III}: The final output is stored directly in GPU memory, ensuring that the data remains within the GPU for immediate access in subsequent SpGEMM cycles.
\end{itemize}

As for GCN systems, in the aggregation step, we first initiate with an analysis of the adjacency matrix's byte size in CSR format and the feature matrix in CSC format, and assess the data compatibility with available GPU and CPU memory capacities, avoiding actual data loading to minimize initial overhead. 


\section{Implementation}
\label{sec:imple}
AIRES is an end-to-end system implemented on top of CUDA v12.2 with 1.5K lines of C++ code.
In our development of AIRES, we implemented block-wise alignment processing on the CPU using C++. 
We employ the NVIDIA GPU Direct Storage (GDS) with the cuFile library v1.7~\cite{gpus_direct_storage} for dynamic scheduling, facilitating rapid data exchanges between storage devices and GPU memory.
Additionally, dynamic memory allocation for DMA transfers was managed using \textit{cudaMalloc}, guided by an analytical model to optimize memory usage. 
This comprehensive approach ensures efficient data handling and computation, leveraging the strengths of both CPU and GPU architectures for out-of-core graph processing.
On top of tiling optimization, we implement a specialized compressed sparse matrix multiplication using CUDA kernels to harness GPU capabilities efficiently.

\section{Evaluation}
\label{sec:eval}

\subsection{Experimental Setup}
\minisection{Hardware.} 
We conduct our experiments on a single machine, configured with a single NVIDIA RTX 4090 GPU with 24\,GB VRAM, and a 13th Gen Intel(R) Core(TM) i9-13900KF CPU with 128\,GB DDR5 memory and 2TB M.2 NVMe SSD.
We model the I/O transfer operations and kernel-level computation latency with simulations, which are profiled using the NVIDIA Nsight system.

\minisection{Model configuration.} We train a GCN network with a feature matrix dimension of 256 with 99\% uniform sparsity ratio, where one training epoch entails multiple cycles of SpGEMM, activation, and backward gradient descent operations over the entire dataset.

\begin{table}[!t]
\centering
\caption{Characteristics of seven diverse graph datasets from SuiteSparse~\cite{davis2011university}, including the number of vertices, edges, the GPU memory requirements, and constraints. }
\label{tab:dataset}
\resizebox{\columnwidth}{!}{
\begin{tabular}{lcccc}
\toprule

\textbf{Dataset} & \begin{tabular}[c]{@{}c@{}}No. Vertices\\ (M)\end{tabular} & \begin{tabular}[c]{@{}c@{}}No. Edges\\ (M)\end{tabular} & \begin{tabular}[c]{@{}c@{}}Memory\\ Req. (GB)\end{tabular} & \begin{tabular}[c]{@{}c@{}}Memory\\ Constraint. (GB)\end{tabular} \\

\midrule
rUSA & 23.94 & 57.70 & 3.31 & 3 \\
kV2a & 55.04 & 117.21 & 6.87 & 6 \\
kU1a & 67.71 & 138.77 & 8.2 & 8 \\ 
socLJ1 & 4.84 & 68.99 & 12.14 & 11 \\ 
kP1a & 139.35 & 297.82 & 17.45 & 16 \\
kA2a & 170.72 & 360.58 & 21.18 & 18 \\
kV1r & 214.00 & 465.41 & 27.18 & 23 \\ 
\bottomrule
\end{tabular}
}
\end{table}

\minisection{Datasets.} 
For evaluation, we adopt a suite of graph datasets from the SuiteSparse matrix collection~\cite{davis2011university}, namely 
road\_usa (rUSA) from street-network datasets, soc-LiveJournal1 (socLJ1) from social-network datasets~\cite{snapnets}, and the rest are from the biomedical GenBank datasets, such as kmer\_V2a (kv2a).
To validate the efficacy of AIRES, we set the GPU memory accordingly to simulate out-of-core scenarios. Memory requirement refers to the combined size of A, B, and C, while memory constraint refers to the total available GPU memory.  
Details of the datasets are shown in Table~\ref{tab:dataset}.

\minisection{Baselines.} 
We compare AIRES with three previous baseline methods and summarized table shown in Table~\ref{tab:Evaluation}:

\begin{itemize}
    \item \textbf{MaxMemory.} MaxMemory represents a naive static method that stores a maximum equal amount of both the adjacency matrix and the feature matrix data in GPU memory, with the remainder stored in CPU memory.
    \item \textbf{UCG~\cite{lin2024unified}.}
    UCG~\cite{lin2024unified} designs a unified CPU-GPU protocol designed to train GCNs by utilizing both CPUs and GPUs collaboratively in multi-GPU settings and dynamically balancing the workload between CPU and GPU.
    \item \textbf{ETC~\cite{gao2024etc}.}
    ETC~\cite{gao2024etc} is the state-of-the-art method batching scheme to allow larger batches while minimizing information loss. 
    It employs a three-step data access policy and an inter-batch pipeline mechanism to reduce redundant data access and minimize CPU-to-GPU data transfer.
    
\end{itemize}

\subsection{Performance Analysis} 
\begin{figure}[!t]
\centerline{\includegraphics[width=0.45\textwidth]{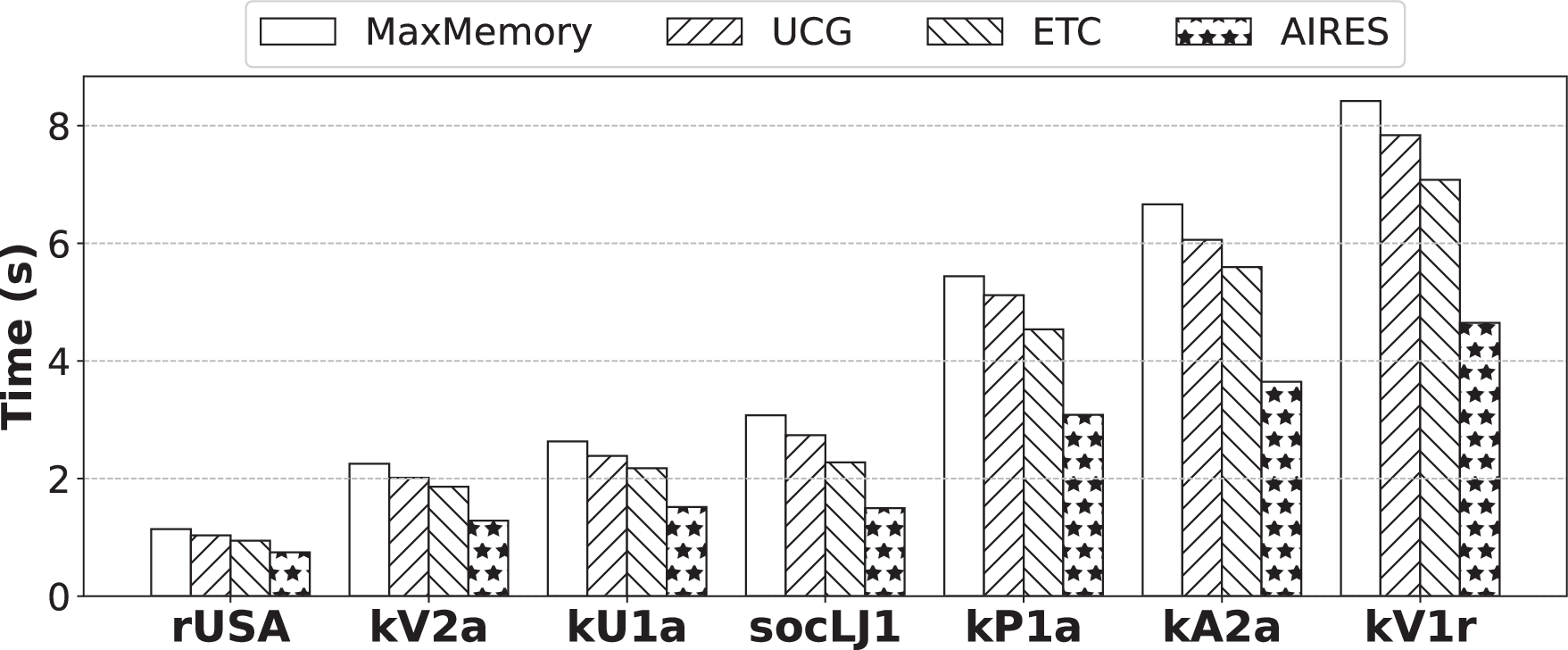}}
\caption{The AIRES speed up over baseline and state-of-the-art for per-epoch execution time for sub-graph.
}
\label{fig:peridot_speedup}
 \end{figure}

\minisection{End-to-End Results.}
Figure~\ref{fig:peridot_speedup} demonstrates the overall performance comparison across five graph datasets.
Due to the limit of computing power, we measure the end-to-end latency every epoch for fair comparison. 
There are three key observations.
First, AIRES achieves consistent speedup overall baseline methods, showing $1.5\sim1.8\times$ over ETC.
Prior works such as UCG are not optimized for out-of-core resource-constrained systems, therefore inducing higher latency and lower throughput.
Second, AIRES is more scalable than previous works.
As the dataset size grows, the speedup of AIRES over MaxMemory
and other methods increases.
Third, AIRES provides consistent latency improvement across datasets of varying sizes.
Specifically, AIRES achieves an average total speedup of 1.8$\times$, 1.7$\times$, and 1.5$\times$ over MaxMemory, UCG, ETC, respectively. 
Such performance gain is attributed to two reasons: 
1) AIRES utilizes a block-wise partitioning strategy to effectively minimize the time spent on merging and staging data;
2) the dynamic scheduling in AIRES features an efficient dual-way data transfer mechanism that leverages both DMA and GDS technologies to further reduce I/O latency.



 \begin{figure*}[!t]
\centerline{\includegraphics[width=0.956\textwidth]{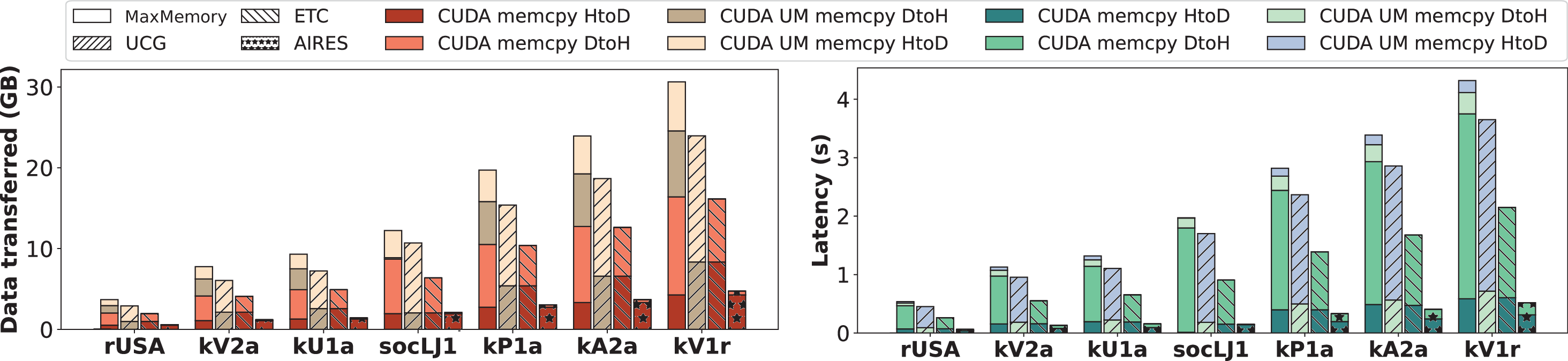}}
\caption{GPU-CPU I/O performance breakdown for different methods. 
On the left is data transferred of CUDA memory operations by dataset and on the right is the average latency measured for each operation measured for per-epoch operation. 
UM stands for Unified Memory, HtoD, and DtoH denote host-to-device and device-to-host, respectively.
}
\label{fig:combination}
\end{figure*}

\begin{figure}[!t]
\vspace{-5mm}
\centerline{\includegraphics[width=0.45\textwidth]{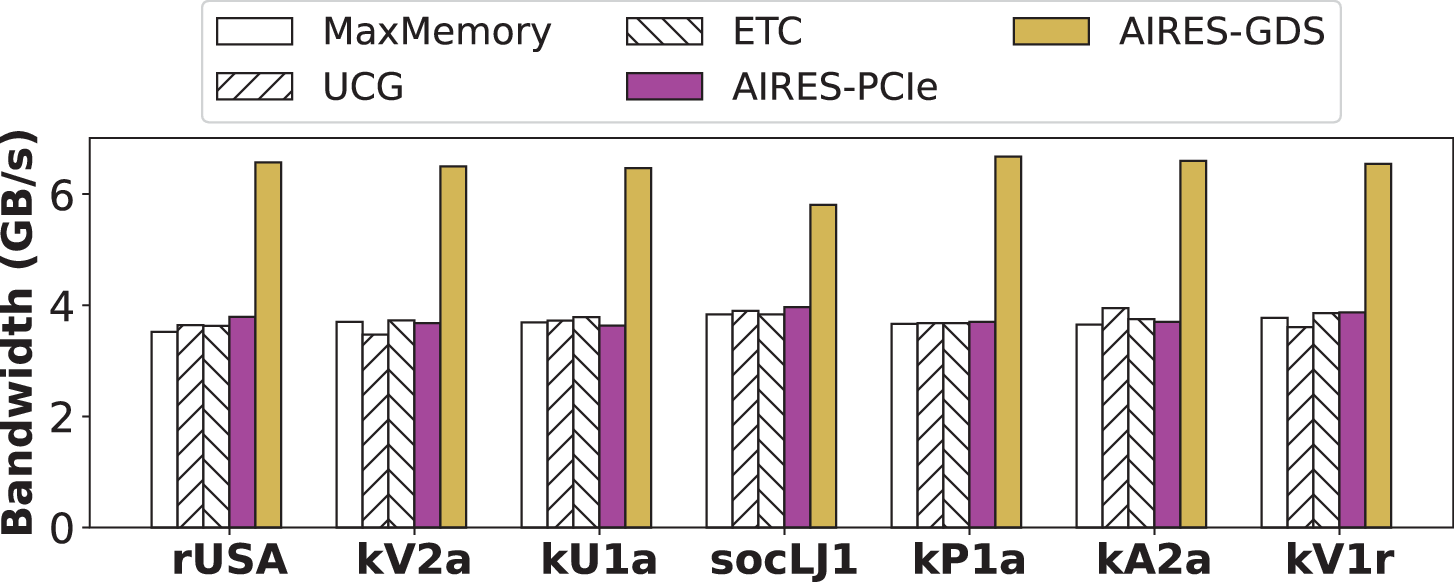}}

\caption{Bandwidth of GPU/CPU-SSD.
GPU-SSD goes through GDS while CPU-SSD is through the PCIe bus.
}
\label{fig:bandwidth}
\end{figure}

\minisection{I/O Performance Breakdown of GPU-CPU.} To better understand the impact of AIRES's block-wise partitioning, we profile the data transfer time and size between GPU and CPU, with results given in Figure~\ref{fig:combination}. 
The GPU I/O operations include CUDA memcpy for host-to-device (HtoD) and device-to-host (DtoH), as well as CUDA unified memory (UM) memcpy operations.
There are three key observations here in Figure~\ref{fig:combination}.
First, AIRES significantly reduces the amount of data transferred between GPU and CPU.
For instance, in kA2a dataset, 
AIRES decreases the amount of data transferred by 84.2\% against the MaxMemory method from 30.4\,GB to 4.83\,GB.
Such improvement is due to two reasons:
1) AIRES removes the repetitive merging overheads in the CPU core by transferring sparse data in a block-wise fashion;
2) AIRES optimizes the memory allocation for the output compress matrix, ensuring that memory usage is minimized while maintaining performance, thereby reducing the overall data that needs to be transferred;
Second, on top of the reduced data transferred, AIRES also provides consistent latency improvement over baseline methods.
For instance, on the kV1r, AIRES reduces the data transfer size and latency by 70\% and 75\%, respectively, compared to the state-of-the-art ETC method at the same time.
Third, AIRES significantly outperforms all other methods across different datasets.
The bigger the dataset size, the larger the latency improvement.

\minisection{I/O Performance Breakdown of GPU/CPU-SSD.}
To better understand the impact of the dual-way data transfer strategy in AIRES, we further perform bandwidth profiling for data movement between GPU/CPU and secondary storage (SSD).
Figure~\ref{fig:bandwidth}, demonstrates that AIRES significantly enhances bandwidth across all datasets when compared to the baseline configuration, thanks to the dual-way path strategy. Specifically, the use of GPU direct storage allows us to transfer more data directly from GPU to SSD instead of the slow PCIe lane, therefore reducing the end-to-end latency.

\begin{table}[!t]
\vspace{5mm}
\caption{Impact of GPU memory constraints on the per-epoch execution time of AIRES and baseline methods. `-' means out-of-memory error.
}
\label{tab:CR_study}
\resizebox{\columnwidth}{!}{
\begin{tabular}{r|c||ccccc}
\toprule
Dataset & \multicolumn{1}{c||}{\begin{tabular}[c]{@{}c@{}}Mem. constraint \\ (GB)\end{tabular}} & MaxMemory & UCG  & ETC & AIRES \\ 
\midrule
\multirow{3}{*}{\textbf{kV1r}} & 24 & 8.47\,s & 7.74\,s & 6.95\,s & \textbf{4.95\,s} \\
 & 21 & - & - & 7.02\,s & \textbf{5.01\,s} \\ 
 &\cellcolor{yellow!20}19 &\cellcolor{yellow!20}- &\cellcolor{yellow!20}- &\cellcolor{yellow!20}- &\cellcolor{yellow!20}\textbf{5.05\,s} \\ \hline
 \multirow{3}{*}{\textbf{kP1a}} & 16 & 5.75\,s & 5.07\,s &4.49\,s & \textbf{3.24\,s} \\
 & 14 & - & - &4.63\,s & \textbf{3.26\,s} \\ 
 &\cellcolor{yellow!20}12 &\cellcolor{yellow!20}- &\cellcolor{yellow!20}- &\cellcolor{yellow!20}- & \cellcolor{yellow!20}\textbf{3.37\,s} \\ \hline
 \multirow{3}{*}{\textbf{socLJ1}} & 11 & 3.18\,s & 2.80\,s & 2.29\,s & \textbf{1.67\,s}  \\
 & 10 & - & - & 2.34\,s & \textbf{1.71\,s} \\ 
 &\cellcolor{yellow!20}8 &\cellcolor{yellow!20}- &\cellcolor{yellow!20}- &\cellcolor{yellow!20}- &\cellcolor{yellow!20}\textbf{1.76\,s} \\ \hline
\end{tabular}
}
\end{table}

\begin{figure}[!t]
\vspace{-5mm}
\centerline{
\includegraphics[width=0.45\textwidth]{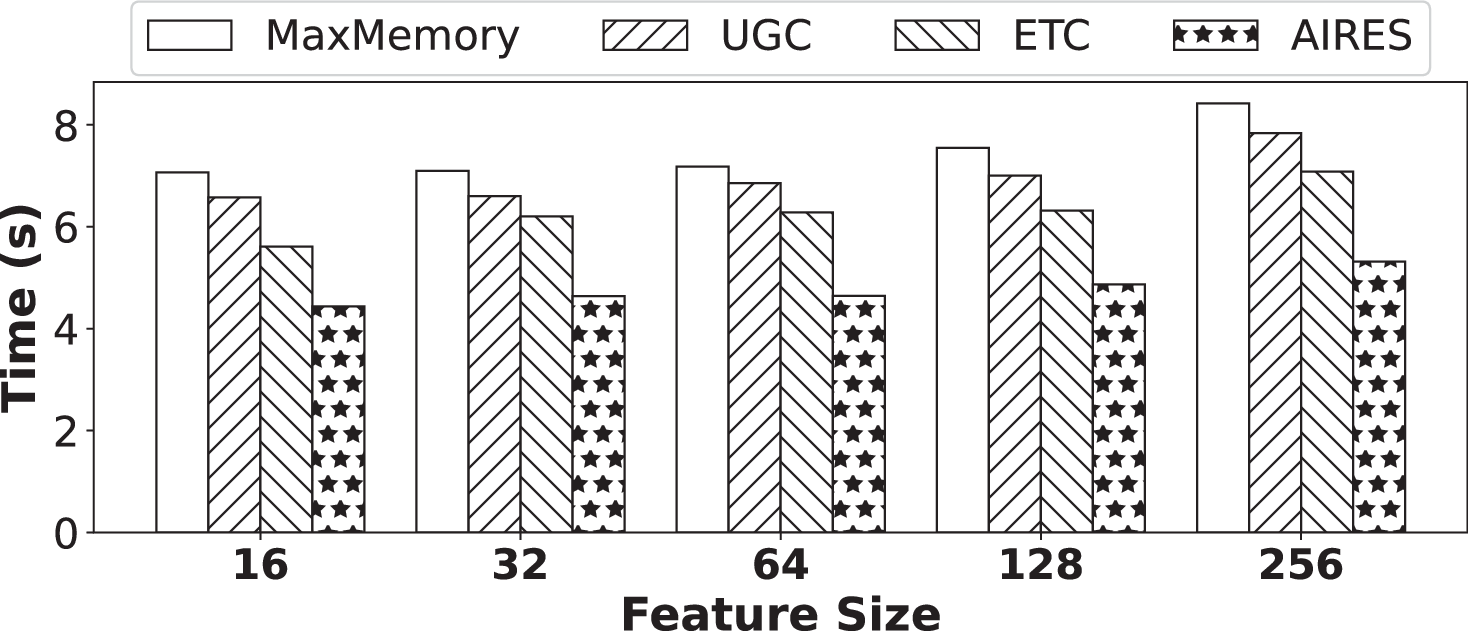}
}

\caption{Per-Epoch execution time evaluation for different GCN feature sizes in AIRES and baselines.
}
\label{fig:different_feature_size}
\end{figure}

\minisection{Ablation Studies.} We further conduct ablation experiments to show the impact of GCN model feature size and GPU memory constraints on the end-to-end per-epoch training latency.
Figure~\ref{fig:different_feature_size} compares the execution time of different methods under varying feature sizes, ranging from 16 to 256.
We can see that AIRES provides consistent speedup across different model configurations, making it a more efficient choice for handling large-scale GCN data.
Table~\ref{tab:CR_study} further demonstrates how different methods perform under varying GPU memory constraints for three datasets.

There are two key observations here. 
First, as the memory constraints decrease, baseline models often run into out-of-core memory (OOM) errors due to the minimum data not available in the GPU memory to complete the computation. 
In contrast, AIRES demonstrates a robust capability to operate effectively with low memory constraints, showcasing its adaptability to stringent memory constraints. 
Second, lower memory constraints correlate with increased computation times, as frequent I/O transfers between CPU memory and GPU HBM become necessary to manage data flow and processing continuity. 
This interaction underscores the importance of dynamic scheduling implemented in AIRES to minimize performance degradation in constrained environments.

\section{Related Work}
\label{sec:related_work}

\minisection{Sparse General Matrix Multiplication.} 
SpGEMM is a fundamental operation in high-performance computing and scientific applications, particularly in solving large sparse systems and graph algorithms~\cite{gao2023systematic}. 
Existing works on SpGEMM have explored various approaches to optimize performance, such as exploiting parallelism~\cite{azad2016exploiting,patwary2015parallel}, minimizing memory usage~\cite{du2022accelerating,chen2020optimizing}, improving load balancing~\cite{anh2016balanced}, and specialized accelerator design~\cite{song2022sextans}. 
Orthogonal to previous works, 
AIRES aims to address the SpGEMM operations in out-of-core scenarios.

\minisection{Memory Optimization.}
Numerous works have attempted to optimize memory allocations for big data applications~\cite{chang2024gmt,lin2024unified}.
Notably, GMT bridges this gap by adopting a customized approach that maximizes the unique properties of GPU memory to ensure that essential data is stored in the fastest accessible memory~\cite{chang2024gmt}.
UCG leverages unified memory to distribute workloads between CPUs and GPUs, instead of offloading the majority of tasks to the GPU as traditional methods do~\cite{lin2024unified}.
In a similar vein, AIRES introduces a dynamic scheduling strategy for memory optimization to perform memory allocations based on computational needs and data characteristics, thereby ensuring optimal performance across various out-of-core computations.



\section{Conclusion}
\label{sec:conclusion}
In this work, we present AIRES, a novel algorithm-system co-design approach dedicated to enhancing the performance of GCNs in resource-constrained systems, such as single GPU-CPU systems.
On the algorithm level, AIRES adopts a row block-wise (RoBW) alignment partitioning method for matrices in compressed format and develops a tiling algorithm for compressed matrix multiplication to facilitate the RoBW.
On the system level, AIRES employs a three-phase dynamic scheduling protocol that features a dual-way data transfer strategy within a tiered memory system to maximize GPU utilization.
Experiments show that AIRES sustains up to 1.8$\times$ latency improvement upon baseline methods.

\section{Acknowledgments}
This work was sponsored in part by the U.S. National Science Foundation (NSF) under Grants 1907765, 2028481, and 2426368.
The authors would like to thank the anonymous ASAP reviewers for their constructive feedback to improve this work.

\bibliographystyle{IEEEtranS}
\bibliography{IEEEabrv,references}

\end{document}